\documentclass[10pt,twocolumn,letterpaper]{article}

\usepackage{cvpr}
\usepackage{times}
\usepackage{epsfig}
\usepackage{graphicx}
\usepackage{amsmath}
\usepackage{amssymb}

\DeclareMathOperator*{\argmax}{argmax}
\usepackage{bbm}

\usepackage[breaklinks=true,bookmarks=false]{hyperref}
\usepackage{tabularx}
\usepackage{booktabs}
\usepackage{multirow}
\usepackage{relsize}

\newcommand*\samethanks[1][\value{footnote}]{\footnotemark[#1]}

\cvprfinalcopy 

\def\cvprPaperID{6348} 

\ifcvprfinal\fi
\begin{document}

\title{Unsupervised learning of action classes with continuous temporal embedding}

\author{Anna Kukleva \thanks{This work was mainly done at University of Bonn. Asterisk denotes equal contribution.}\\
University of Bonn\\
Germany\\
{\tt\small s6ankukl@uni-bonn.de}
\and
Hilde Kuehne \samethanks\\
MIT-IBM Watson Lab \\
Cambridge, MA\\
{\tt\small kuehne@ibm.com}
\and
Fadime Sener, Juergen Gall\\
University of Bonn\\
Germany\\
{\tt\small sener,gall@iai.uni-bonn.de}
}

\maketitle

\begin{abstract}
The task of temporally detecting and segmenting actions in untrimmed
videos has seen an increased attention recently. 
One problem in this context arises from the need to define and label action boundaries to create annotations for training which is very time and cost intensive.
To address this issue, we propose an unsupervised approach for learning action classes from untrimmed video sequences. To this end, we use a continuous temporal embedding of framewise features to benefit from the sequential nature of activities. Based on the latent space created by the embedding, we identify clusters of temporal segments across  all videos that correspond to semantic meaningful action classes.
The approach is evaluated on three challenging datasets, namely the Breakfast dataset, YouTube Instructions, and the 50Salads dataset. While previous works assumed that the videos contain the same high level activity, we furthermore show that the proposed approach can also be applied to a more general setting where the content of the videos is unknown.
\end{abstract}
\vspace{-0.5cm}
\section{Introduction}
The task of action recognition has seen tremendous success over the last years. So far, high-performing approaches require full supervision for training. But acquiring frame-level annotations of actions in untrimmed videos is very expensive and impractical for very large datasets.     
Recent works, therefore, explore alternative ways of training action recognition approaches without having full frame annotations at training time. Most of those concepts, which are referred to as weakly supervised learning, rely on ordered action sequences which are given for each video in the training set.

Acquiring ordered action lists, however, can also be very time consuming and it assumes that it is already known what actions are present in the videos before starting the annotation process. For some applications like indexing large video datasets or human behavior analysis in neuroscience or medicine, it is often unclear what action should be annotated. It is therefore important to discover actions in large video datasets before deciding which actions are relevant or not. 
Recent works~\cite{sener2018unsupervised, alayrac2016unsupervised} therefore proposed the task of unsupervised learning of actions in long, untrimmed video sequences. Here, only the videos themselves are used
and the goal is to identify clusters of temporal segments across all videos that correspond to semantic meaningful action classes.

In this work we propose a new method for unsupervised learning of actions from long video sequences, which is based on the following contributions. The first contribution is the learning of a continuous temporal embedding of frame-based features. The embedding exploits the fact that some actions need to be performed in a certain order and we use a network to learn an embedding of frame-based features with respect to their relative time in the video. As the second contribution, we propose a decoding of the videos into coherent action segments based on an ordered clustering of the embedded frame-wise video features. To this end, we first compute the order of the clusters with respect to their timestamp. Then a Viterbi decoding approach is used such as in~\cite{richard2018nnviterbi, vetibi_koller_2017, richard2017weakly, laxton_viterbi_2007} which maintains an estimate of the most likely activity state given the predefined order.  

We evaluate our approach on the Breakfast~\cite{kuehne2014language} and YouTube Instructions datasets \cite{alayrac2016unsupervised}, following the evaluation protocols used in \cite{sener2018unsupervised, alayrac2016unsupervised}. We also conduct experiments on the 50Salads dataset~\cite{50salads} where the videos are longer and contain more action classes.
Our approach outperforms the state-of-the-art in unsupervised learning of action classes from untrimmed videos by a large margin. The evaluation protocol used in previous works, however, divides the datasets into distinct clusters of videos using the ground-truth activity label of each video, \ie, unsupervised learning and evaluation are performed only on videos, which contain the same high level activity. This simplifies the problem since in this case most of the actions occur in all videos. 

As a third contribution, we therefore propose an extension of our approach that allows to go beyond the scenario of processing only videos from known activity classes, \ie, we discover semantic action classes from all videos of each dataset at once, in a completely unsupervised way without any knowledge of the related activity. To this end, we learn a continuous temporal embedding for all videos and use the embedding to build a representation for each untrimmed video. After clustering the videos, we identify consistent video segments for all videos within a cluster. In our experiments, we show that the proposed approach not only outperforms the state-of-the-art using the simplified protocol, but it is also capable to learn actions in a completely unsupervised way. Code is available on-line.\footnote{\url{https://github.com/annusha/unsup\_temp\_embed}}

\section{Related work}

Action recognition~\cite{laptev08learning,wang2013action,simonyan2014two,carreira2017quo,Diba} as well as the understanding complex activities~\cite{kuehne2014language, yeung2016end, shou2017cdc} has been studied for many years with a focus on fully supervised learning. Recently, there has been an increased interest in methods that can be trained with less supervision. One of the first works in this field has been proposed by Laptev~\etal \cite{laptev08learning} where the authors learn actions from movie scripts. 
Another dataset that follows the idea of using subtitles has been proposed by Alayrac~\etal \cite{alayrac2016unsupervised}, also using YouTube videos to automatically learn actions from instructional videos. A multi-modal version of this idea has been proposed by~\cite{malmaud15what}. Here, the authors also collected cooking videos from YouTube and used a combination of subtitles, audio, and vision to identify receipt steps in videos. Another way of learning from subtitles is proposed by Sener~\etal \cite{senerozan2015unsupervised} by representing each frame via the occurrence of actions atoms given the visual comments at this point.
These works, however, assume that the narrative text is well-aligned with the visual data. Another form of weak supervision are video transcripts~\cite{huang2016connectionist,kuehne2017weakly,richard2017weakly,ding2017weakly,richard2018nnviterbi}, which provide the order of actions but that are not aligned with the videos, or video tags~\cite{WangXLG17,richard2017temporal}.

There are also efforts for unsupervised learning of action classes. One of the first works that was tackling the problem of human motion segmentation without training data was proposed by Guerra-Filho and Aloimonos~\cite{Guerra2007Language}. They propose a basic segmentation with subsequent clustering based on sensory-motor data. Based on those representations, they propose the application of a parallel synchronous grammar system to learn atomic action representations similar to words in language. Another work in this context is proposed by Fox~\etal \cite{Fox2013Joint} where a Bayesian nonparametric approach helps to jointly model multiple related time series without further supervision. They apply their work on motion capture data.

In the context of video data, the temporal structure of video data has been exploited to fine-tune networks on training data without labels~\cite{WangG15,buechler2017LSTM}. The temporal ordering of video frames has also been used to learn feature representations for action recognition~\cite{OPN, Ramanathan_2015_ICCV, Fernando_2015, cherian2017generalized}. 
Lee \etal \cite{OPN} learn a video representation in an unsupervised manner by solving a sequence sorting problem. 
Ramanathan \etal \cite{Ramanathan_2015_ICCV} build their temporal embedding by leveraging contextual information of each frame on different resolution levels. Fernando \etal \cite{Fernando_2015} presented a method to capture the temporal evolution of actions based on frame appearance by learning a frame ranking function per video. In this way, they obtain a compact latent space for each video separately. A similar approach to learn a structured representation of postures and their temporal development was proposed by Milbich \etal \cite{milbich2017unsupervised}.  
While these approaches address different tasks, Sener \etal \cite{sener2018unsupervised} proposed an unsupervised approach for learning action classes. They introduced an iterative approach which alternates between discriminative learning of the appearance of sub-activities from visual features and generative modeling of the temporal structure of sub-activities using a Generalized Mallows Model.

\begin{figure*}[t]
\begin{center}
    \includegraphics[width=0.99\linewidth]{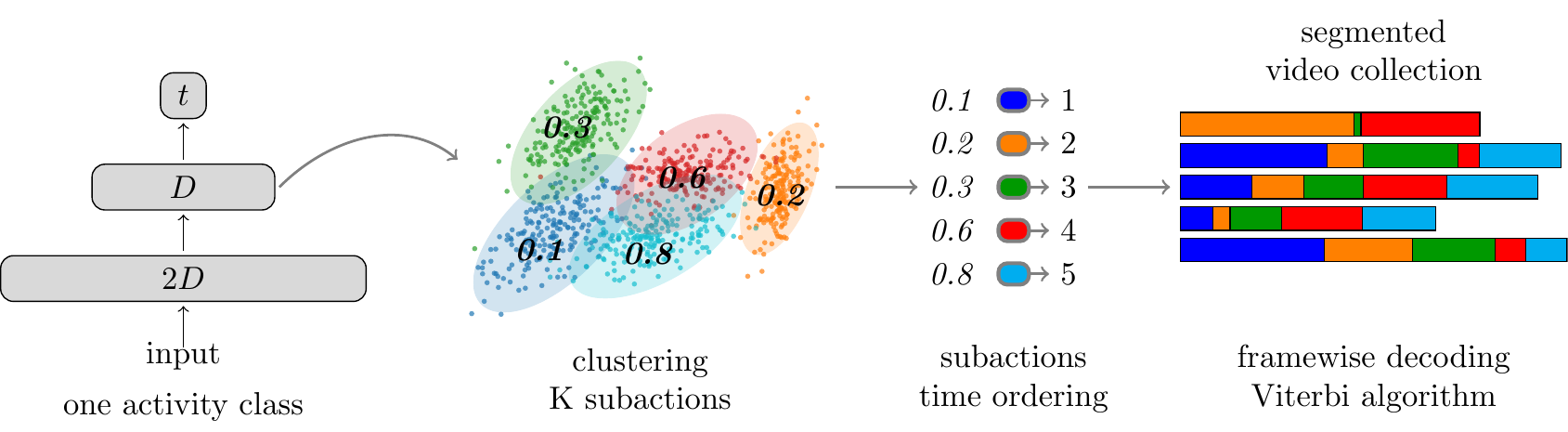}
\end{center}
   \caption{Overview of the proposed pipeline. We first compute the embedding of features with respect to their relative time stamp. The resulting embedded features are then clustered and the mean temporal appearance of each cluster is computed and the ordering of clusters is computed. Each video is then decoded with respect to this ordering given the overall proximity of each frame to each cluster. }
\label{fig:local}
\end{figure*}

\section{Unsupervised Learning of Action Classes}
\subsection{Overview}

As input we are given a set $\{ \boldsymbol{X_m}\}_{m=1}^M$ of $M$ videos and each video $\boldsymbol{X_m}=\{x_{mn}\}_{n=1}^{N_m}$ is represented by $N_m$ framewise features. The task is then to estimate the subaction label $l_{mn} \in \{1, \ldots, K\}$ for each video frame $x_{mn}$. Following the protocol of~\cite{alayrac2016unsupervised,sener2018unsupervised}, we define the number of possible subactions $K$ separately for each activity as the maximum number of possible subactions as they occur in the ground-truth. The values of $K$ are provided in the supplementary material.

Fig.~\ref{fig:local} provides an overview of our approach for unsupervised learning of actions from long video sequences. First, we learn an embedding of all features with respect to their relative time stamp as described in Sec.~\ref{sec:Continuous_Temporal_Embedding}. The resulting embedded features are then clustered and the mean temporal occurrence of each cluster is computed. This step, as well as the temporal ordering of the clusters is described in Sec.~\ref{sec:Clustering_and_ordering}. Each video is then decoded with respect to this ordering given the overall proximity of each frame to each cluster as described in Sec.~\ref{sec:Viterbi_Decoding}.

We also present an extension to a more general protocol, where the videos have a higher diversity. Instead of assuming as in~\cite{alayrac2016unsupervised,sener2018unsupervised} that the videos contain the same high-level activity, we discuss the completely unsupervised case in Sec.~\ref{sec:Unsupervised_unknown_activities}. We finally introduce a background model to address background segments in Sec.~\ref{sec:Background Model}.

\subsection{Continuous Temporal Embedding}
\label{sec:Continuous_Temporal_Embedding}

The idea of learning a continuous temporal embedding relies on the assumption that similar subactions tend to appear at a similar temporal range within a complex activity. For instance a subaction like ``take cup'' will usually occur in the beginning of the activity ``making coffee''. After that people probably pour coffee into the mug and finally stir coffee. Thus many subactions that are executed to conduct a specific activity are softly bound to their temporal position within the video. 

To capture the combination of visual appearance and temporal consistency, we model a continuous latent space by capturing simultaneously relative time dependencies and the visual representation of the frames. 
For the embedding, we train a network architecture which optimizes the embedding of all framewise features of an activity with respect to their relative time $t(x_{mn}) = \frac{n}{N_m}$.
As shown in Fig.~\ref{fig:local}, we take an MLP with two hidden layers with dimensionality $2D$ and $D$, respectively, and logistic activation functions. As loss, we use the mean squared error between the predicted time stamp and the true time stamp $t(x_{mn})$ of the feature. The embedding is then given by the second hidden layer. 

Note that this embedding does not use any subaction label associations as in~\cite{sener2018unsupervised,alayrac2016unsupervised}, thus the network needs to be trained only once instead of retraining the model at each iteration. 
For the rest of the paper, $x_{mn}$ denotes the embedded $D$-dimensional features.   

\begin{figure*}[t]
\begin{center}
    \includegraphics[width=0.98\linewidth]{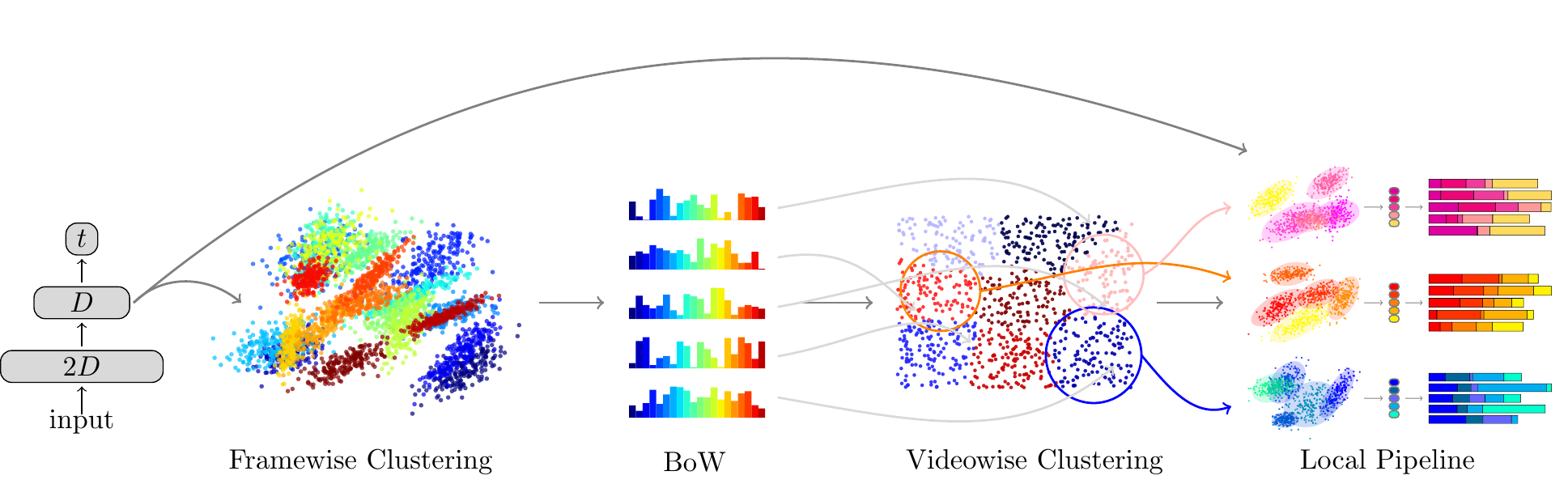}
\end{center}
   \caption{Proposed pipeline for unsupervised learning with unknown activity classes. We first compute an embedding with respect to the whole dataset at once. In a second step, features are clustered in the embedding space to build a  bag-of-words representation for each video. We then cluster all videowise vectors into $K'$ clusters and apply the previously described method for each video set.}
\label{fig:global}
\end{figure*}

\subsection{Clustering and Ordering}
\label{sec:Clustering_and_ordering}

After the embedding, the features of all videos are clustered into $K$ clusters by k-Means. 
Since in Sec.~\ref{sec:Viterbi_Decoding} we need the probability $p(x_{mn}| k)$, \ie, the probability that the embedded feature $x_{mn}$ belongs to cluster $k$, we estimate a $D$-dimensional Gaussian distribution for each cluster:  
\begin{equation}
    p(x_{mn}| k) = \mathcal{N}(x_{mn}; \mu_{k}, \Sigma_{k}). 
\label{eq:frame_GMM_prob}
\end{equation}
Note that this clustering does not define any specific ordering. To order clusters with respect to their temporal occurrence, we compute the mean over time stamps of all frames belonging to each cluster
\begin{equation}
\begin{split}
    &X(k) = \{x_{mn} | p(x_{mn} | k) \geq p(x_{mn} | k'), \forall k' \neq k\}, \\
    &t(k) = \frac{1}{|X(k)|}\sum_{x_{mn} \in X(k)}t(x_{mn}). \\
\end{split}
\end{equation}

The clusters are then ordered with respect to the time stamp so that $\{k_1, .., k_K\}$ is the set of ordered cluster labels subject to $0 \leq t({k}_1) \leq .. \leq t({k}_K) \leq 1$.  
The resulting ordering is then used for the decoding of each video.

\subsection{Frame Labeling}
\label{sec:Viterbi_Decoding}

We finally temporally segment each video $\boldsymbol{X_m}$ separately, \ie, we assign each frame $x_{mn}$ to one of the ordered clusters $l_{mn} \in \{k_1, \ldots, k_K\}$. We first calculate the probability of each frame that it belongs to cluster $k$ as defined by~\eqref{eq:frame_GMM_prob}.
Based on the cluster probabilities for the given video, we want to maximize the probability of the sequence following the order of the clusters ${k}_1 \rightarrow .. \rightarrow {k}_K$  to get consistent assignments for each frame of the video:

\begin{align}   \label{probAlignment}
    \hat{l}_{1}^{N_m} =& \argmax_{{l}_{1}, .., {l}_{N_m}} p\big(x_{1}^{N_m} | {l}_{1}^{N_m}\big)  \\
    =& \argmax_{{l}_1, .., {l}_{N_m}} \prod_{n=1}^{N_m} p\big(x_{mn} | {l}_{n}\big) \cdot p\big({l}_{n} | {l}_{n-1}\big)\nonumber,
\end{align}
where $ p(x_{mn}|l_n=k) $ is the probability that $ x_{mn} $ belongs to the cluster $ {k}$, and 
$ p({l}_{n} | {l}_{n-1})$ are the 
transition probabilities of moving from the label ${l}_{n-1}$ at frame $n-1$ to the next label ${l}_{n}$ at frame $n$,
\begin{align}
    p(l_n | {l}_{n-1}) = \mathbbm{1}_{0 \leq l_{n} - l_{n-1} \leq 1}. 
\end{align}
This means that we allow either a transition to the next cluster in the ordered cluster list or we keep the cluster assignment of the previous frame.  
Note that \eqref{probAlignment} can be solved efficiently using a Viterbi algorithm.

\subsection{Unknown Activity Classes}
\label{sec:Unsupervised_unknown_activities}

So far we discussed the case of applying unsupervised learning to a set of videos that all belong to the same activity. When moving to a larger set of videos without any knowledge of the activity class, the assumption of sharing the same subactions within the collection cannot be applied anymore. As it is illustrated in Fig.~\ref{fig:global}, we therefore cluster the videos first into more consistent video subsets.

Similar to the previous setting, we learn a $D$-dimensional embedding of the features but the embedding is not restricted to a subset of the training data, but it is computed for the whole dataset at once. 
Afterward, the embedded features are clustered in this space to build a video representation based on bag-of-words using quantization with a soft assignment. In this way, we obtain a single bag-of-words feature vector per video sequence. Using this representation, we cluster the videos into $K'$ video sets. For each video set, we then separately infer clusters for subactions and assign them to each video frame as in Fig.~\ref{fig:local}. However, we do not learn an embedding for each video set but use the embedding learned on the entire dataset for each video set. The impact of $K$ and $K'$ will be evaluated in the experimental section.

\subsection{Background Model}
\label{sec:Background Model}

As subactions are not always executed continuously and without interruption, we also address the problem of modeling a background class.
In order to decide if a frame should be assigned to one of the $K$ clusters or the background, we introduce a parameter $\tau$ which defines the percentage of features that should be assigned to the background. To this end, we keep only $1-\tau$ percent of the points within each cluster that are closest to the cluster center and add the other features to the background class. For the labeling described in Sec.~\ref{sec:Viterbi_Decoding}, we remove all frames that have been already assigned to the background before estimating $l_{mn} \in \{k_1, \ldots, k_K\}$ \eqref{probAlignment}, \ie, the background frames are first labeled and the remaining frames are then assigned to the ordered clusters $\{k_1, \ldots, k_K\}$.

\section{Evaluation}

\subsection{Dataset}

We evaluate the proposed approach on three challenging datasets: Breakfast~\cite{kuehne2014language}, YouTube Instructional~\cite{alayrac2016unsupervised}, and 50Salads~\cite{50salads}.  

The Breakfast dataset is a large-scale dataset that comprises ten different complex activities of performing common kitchen activities with approximately eight subactions per activity class. 
The duration of the videos varies significantly, \eg \textit{coffee} has an average duration of 30 seconds while cooking \textit{pancake} takes roughly 5 minutes. 
Also in regards to the subactivity ordering, there are considerable variations. 
For evaluation, we use reduced Fisher Vector features as proposed by~\cite{kuehne2016end} and used in~\cite{sener2018unsupervised} and we follow the protocol of~\cite{sener2018unsupervised}, if not mentioned otherwise.

The YouTube Instructions dataset contains 150 videos from YouTube with an average length of about two minutes per video. There are five primary tasks: \textit{making coffee, changing car tire, cpr, jumping car, potting a plant}. The main difference with respect to the Breakfast dataset is the presence of a background class. The fraction of background within different tasks varies from 46\% to 83\%. We use the original precomputed features provided by~\cite{alayrac2016unsupervised} and used by~\cite{sener2018unsupervised}. 

The 50Salads dataset contains 4.5 hours of different people performing a single complex activity,  making mixed salad. Compared to the other datasets, the videos are much longer with an average video length of 10k frames. 
We perform evaluation on two different action granularity levels proposed by the authors: mid-level with 17 subaction classes and eval-level with 9 subaction classes.
\subsection{Evaluation Metrics}

Since the output of the model consists of temporal subaction bounds without any particular correspondences to ground-truth labels, we need a one-to-one mapping between $\{k_1, .., k_K\}$ and the $K$ ground-truth labels to evaluate and compare the method. Following~\cite{sener2018unsupervised} and \cite{alayrac2016unsupervised}, we use the Hungarian algorithm to get a one-to-one matching and report accuracy as the mean over frames (MoF) for the Breakfast and 50Salads datasets.
Note that especially MoF is not always suitable for imbalanced datasets. We therefore also report the Jaccard index as intersection over union (IoU) as an additional measurement.
For the YouTube Instruction dataset, we also report the F1-score since it is used in previous works. Precision and recall are computed by evaluating if the time interval of a segmentation falls inside the corresponding ground-truth interval. To check if a segmentation matches a time interval, we randomly draw $15$ frames of the segments. The detection is considered correct if at least half of the frames match the respective class, and incorrect otherwise. Precision and recall are computed for all videos and F1 score is computed as the harmonic mean of precision and recall.

\subsection{Continuous Temporal Embedding}
\label{sec:exp_temp_embed}
In the following, we first evaluate our approach for the case of know activity classes to compare with~\cite{sener2018unsupervised} and~\cite{alayrac2016unsupervised} and consider the case of completely  unsupervised learning in Sec.~\ref{sec:eval_unsupervised}. First, we analyze the impact of the proposed temporal embedding by comparing the proposed method to other embedding strategies as well as to different feature types without embedding on the Breakfast dataset. As features we consider AlexNet fc6 features, pre-trained on ImageNet as used in \cite{Ramanathan_2015_ICCV}, I3D features~\cite{carreira2017quo} based on the RGB and flow pipeline, and pre-computed dense trajectories~\cite{wang2013action}. We further compare with previous works with a focus on learning the temporal embedding~\cite{Ramanathan_2015_ICCV, Fernando_2015}. We trained these models following the settings of each paper and construct the latent space, which is used to substitute ours. 

As can be seen in Table~\ref{tab:eval_temp_embedding}, the results with the continuous temporal embedding are clearly outperforming all the above-mentioned approaches with and without temporal embedding. We also used OPN~\cite{OPN} to learn an embedding, which is then used in our approach. However, we observed that for long videos nearly all frames where assigned to a single cluster. When we exclude the long videos with degenerated results, the MoF was lower compared to our approach.

\begin{table}[t] \footnotesize
   \centering
   \begin{tabularx}{0.45\textwidth}{lXc}
        \toprule
        \multicolumn{3}{c}{Comp. of temporal embedding strategies}\\
        \cmidrule(lr){1-3}           
        \cmidrule(lr){1-3}
        \textit{ImageNet \cite{imagenet_NIPS2012} + proposed}  &  & $  21.2\%  $  \\
        \textit{I3D \cite{carreira2017quo} + proposed}  &  & $  25.1\%  $  \\
        \textit{dense trajectories~\cite{wang2013action} + proposed}  &  & $  31.6\%  $  \\
        \cmidrule(lr){1-3}   
        \textit{video vector \cite{Ramanathan_2015_ICCV} + proposed}  &  & $ 30.1\% $  \\
        \textit{video darwin \cite{Fernando_2015} + proposed}  &  & $ 36.6\% $  \\
        \textit{ours}  &  & $ 41.8\% $  \\
        \bottomrule
    \end{tabularx}
    \vspace{2mm}
    \caption{ Evaluation of the influence of the temporal embedding. Results are reported as MoF accuracy on the Breakfast dataset. } 
    \label{tab:eval_temp_embedding}
\end{table}

\subsection{Mallow vs.\ Viterbi}
\label{sec:eval_inference}

We compare our approach, which uses Viterbi decoding, with the Mallow model decoding that has been proposed in \cite{sener2018unsupervised}. The authors propose a rankloss embedding over all video frames from the same activity with respect to a pseudo ground-truth subaction annotation. The embedded frames of the whole activity set are then clustered and the likelihood for each frame and for each cluster is computed. For the decoding, the authors build a histogram of features with respect to their clusters with a hard assignment and set the length of each action with respect to the overall amount of features per bin. After that, they apply a Mallow model to sample different orderings for each video with respect to the sampled distribution. The resulting model is a combination of Mallow model sampling and action length estimation based on the frame distribution.

\begin{figure}[t]
\begin{center}
    \includegraphics[width=0.85\linewidth]{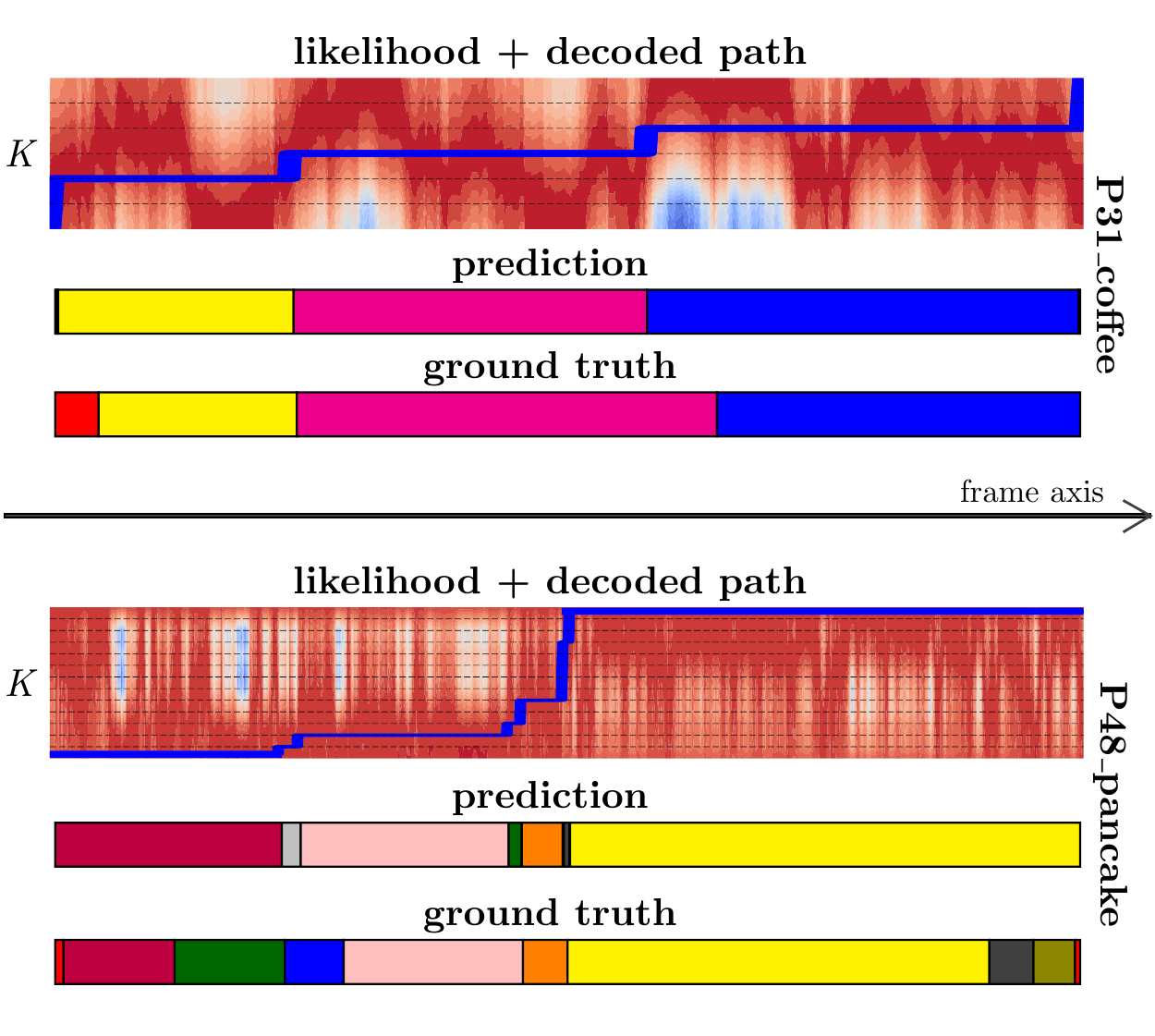}
\end{center}
  \caption{Comparison of Viterbi decoded paths with respective predicted and ground-truth segmentation for two videos. The observation probabilities with red indicating high and blue indicating low probabilities of belonging to subactions. It shows that the decoding assigns most frames to occurring subactions while marginalizing actions that do not occur in the sequence by assigning only a few frames.}
\label{fig:exp_Viterbi_decoding}
\end{figure}

For the first experiment, we evaluated the impact of the different decoding strategies with respect to the proposed embedding. In Table \ref{tab:eval_Inference_strategy} we compare the results of decoding with the Mallow model only, Viterbi only, and a combination of Mallow-Viterbi decoding. For the combination, we first sample the Mallow ordering as described by \cite{sener2018unsupervised} leading to an alternative ordering. We then apply a Viterbi decoding to the new as well as to the original ordering and choose the sequence with the higher probability. It shows that the original combination of Mallow model and multinomial distribution sampling performs worst on the temporal embedding. Also, the combination of Viterbi and Mallow model can not outperform the Viterbi decoding alone. To have a closer look, we visualize the observation probabilities as well as the resulting decoding path over time for two videos in Fig.~\ref{fig:exp_Viterbi_decoding}. It shows that the decoding, which is always given the full sequence of subactions, is able to marginalize subactions that do not occur in the video by just assigning only very few frames to those ones and the majority of frames to the clusters that occur in the video. This means that the effect of marginalization allows to discard subactions that do not occur. Overall, it turns out that this strategy of marginalization usually performs better than re-ordering the resulting subaction sequence as done by the Mallow model.
To further compare the proposed setup to \cite{sener2018unsupervised}, we additionally compare the impact of different decoding strategies, Mallow model and Viterbi, with respect to the two embeddings, rankloss~\cite{sener2018unsupervised} and continuous temporal embedding, in Table \ref{tab:eval_Rankloss_temp_embedding}. It shows that the rankloss embedding works well in combination with the multinomial Mallow model, but fails when combined with Viterbi decoding because of the missing temporal prior in this case, whereas the Mallow model is not able to decode sequences in the continuous temporal embedding space. This shows the necessity of a suitable combination of both, the embedding and the decoding strategy.  

\begin{table}[t] \footnotesize
   \centering
   \begin{tabularx}{0.45\textwidth}{lXc}
        \toprule
        \multicolumn{3}{c}{ Mallow vs.\ Viterbi }\\
        \cmidrule(lr){1-3}           
        & & Acc. (MoF)  \\
        \cmidrule(lr){1-3}           
        \textit{Mallow+multi only}    & &  $ 29.5\%  $  \\
        \textit{Mallow-Viterbi}    & & $  34.8\%  $    \\
        \textit{Viterbi only}    & &  $ 41.8\% $   \\
        \bottomrule
    \end{tabularx}
    \vspace{2mm}
    \caption{Comparison of the Mallow model and Viterbi decoding. Results are reported as MoF accuracy on the Breakfast dataset.  } 
    \label{tab:eval_Inference_strategy}
\end{table}

 \begin{table}[t] \footnotesize
   \centering
   \begin{tabularx}{0.45\textwidth}{lXcXc}
        \toprule
        \multicolumn{5}{c}{Comparison with rankloss and Mallow model}\\
        \cmidrule(lr){1-5}           
        & & Rankloss emb.  & & Temp. emb.  \\
        \cmidrule(lr){1-5}           
        \textit{Mallow model (MoF)}    & & $  34.6\%  $    & & $  29.5\%  $    \\
        \textit{Viterbi dec. (MoF)}    & &  $ 27.1\% $   & &  $ 41.8\% $   \\
        \bottomrule
    \end{tabularx}
    \vspace{2mm}
    \caption{Comparison of proposed embedding and Viterbi decoding with respect to the previously proposed Mallow model \cite{sener2018unsupervised}. Results are reported as MoF accuracy on the Breakfast dataset.  } 
    \label{tab:eval_Rankloss_temp_embedding}
\end{table}
\subsection{Background Model}
\label{sec:eval_background}
\begin{figure}[t]
\begin{center}
    \includegraphics[width=0.9\linewidth]{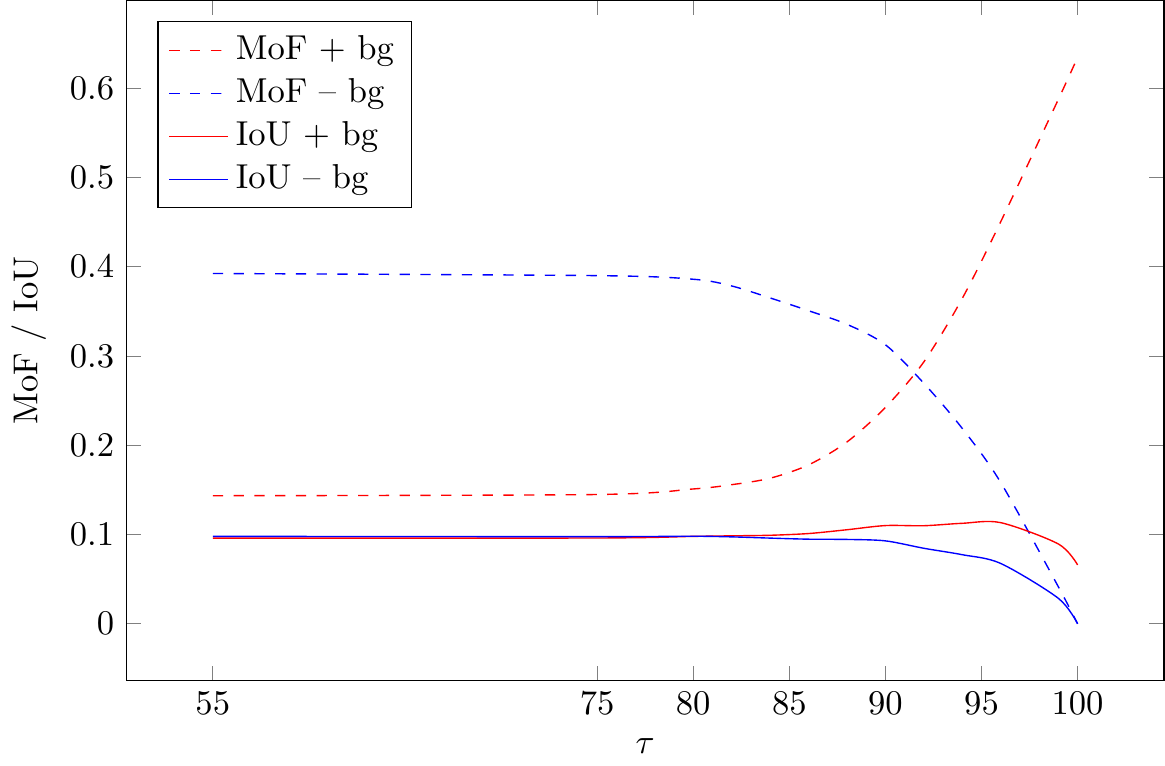}
\end{center}
   \caption{Evaluation of different accuracy measurements with respect to the amount of sampled background on the YouTube Instructions dataset. }
\label{fig:eval_background_YTI}
\end{figure}
Finally, we assess the impact of the proposed background model for the given setting. For this evaluation, we choose the YouTube Instructions dataset. Note that for this dataset, two different evaluation protocols have been proposed so far. \cite{alayrac2016unsupervised} evaluates results on the YTI dataset usually without any background frames, which means that during evaluation, only frames with a class label are considered and all background frames are ignored. Note that in this case it is not penalized if estimated subactions become very long and cover the background. Including background for a dataset with a high background portion, however, leads to the problem that a high MoF accuracy is achieved by labeling most frames as background. 
We therefore introduce for this evaluation the Jaccard index as intersection over union (IoU) as additional measurement, which is also common in comparable weak learning scenarios \cite{richard2017weakly}. 
For the following evaluation, we vary the ratio of desired background frames as described in Sec.~\ref{sec:Background Model} from $75\%$ to $99\%$ and show the results in Fig.~\ref{fig:eval_background_YTI}. As can be seen, a smaller background ratio leads to better results when computing MoF without background, whereas a higher background ratio leads to better results when the background is considered in the evaluation. When we compare it to the IoU with and without background, it shows that the IoU without background suffers from the same problems as the MoF in this case, but that the IoU with background gives a good measure considering the trade-off between background and class labels. For $\tau$ of $75 \%$, our approach achieves $9.6\%$ and $9.8\%$ IoU with and without background, respectively, and $14.5\%$ and $39.0\%$ MoF with and without background, respectively. 
\begin{figure}[t]
\begin{center}
    {\includegraphics[width=0.45\linewidth]{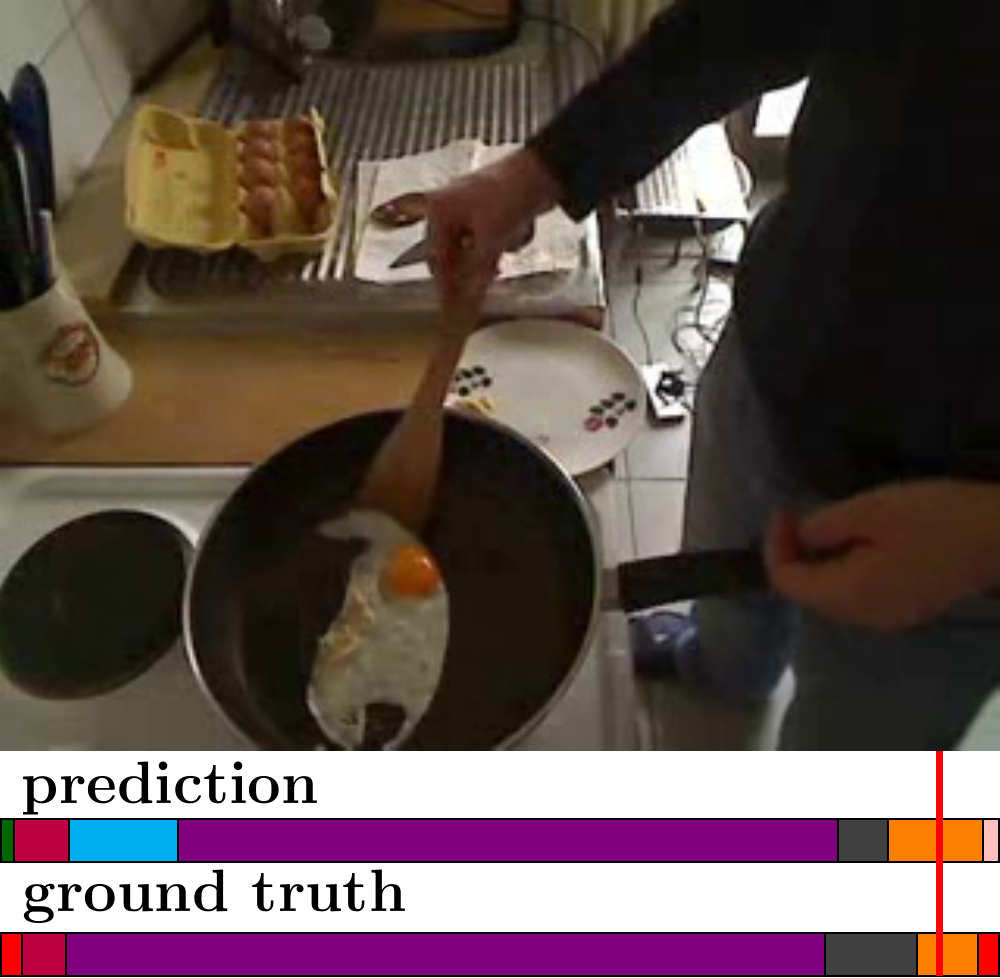}}
    {\includegraphics[width=0.45\linewidth]{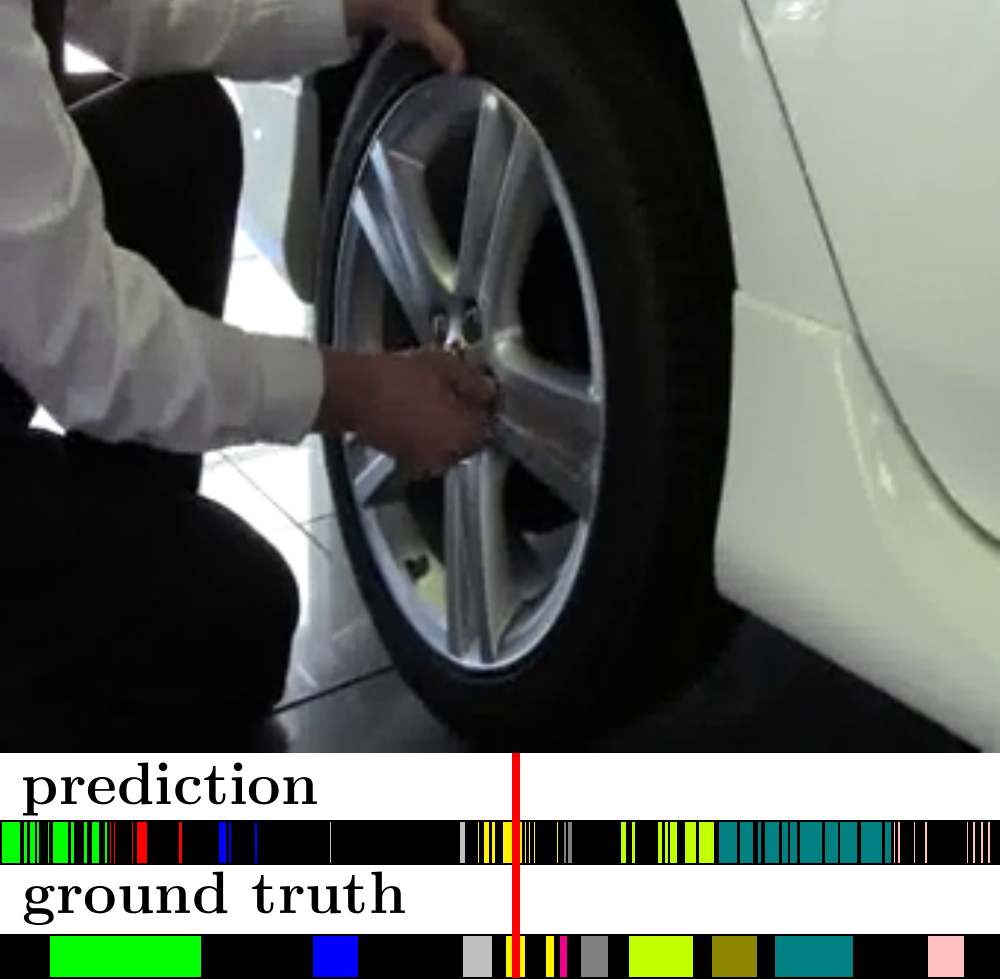}}
\end{center}
\caption{Qualitative analysis of segmentation results on the Breakfast and the YouTube Instructions dataset. }
\label{fig:segm_frames}
\end{figure}
%
\subsection{Comparison to State-of-the-art}
 \begin{table}[t] \footnotesize
   \centering
   \begin{tabularx}{0.45\textwidth}{lXcXc}
        \toprule
        \multicolumn{5}{c}{ Breakfast dataset}\\
        \cmidrule(lr){1-5}           
        & \multicolumn{4}{c}{Fully supervised}\\
        \cmidrule(lr){1-5}           
        & & & &  MoF   \\
        \cmidrule(lr){1-5}           
        HOGHOF+HTK \cite{kuehne2014language}    & &  & &  $ 28.8\% $  \\
        TCFPN \cite{ding2017weakly}    & &  & &  $ 52.0\% $  \\
        HTK+DTF w.\ PCA \cite{kuehne2016end}    & &  & &  $ 56.3\% $  \\
        RNN+HMM \cite{ding2017weakly}    & &  & &  $ 60.6\% $  \\
        \cmidrule(lr){1-5}           
        & \multicolumn{4}{c}{Weakly supervised}\\
        \cmidrule(lr){1-5}           
        & &  & & MoF   \\
        \cmidrule(lr){1-5}           
        ECTC \cite{huang2016connectionist}   & &  & &  $ 27.7\% $   \\
        GMM+CNN \cite{kuehne2017weakly}   & &  & &  $ 28.2\% $   \\
        RNN-FC \cite{richard2017weakly}   & &  & &  $ 33.3\% $   \\
        TCFPN \cite{ding2017weakly}   & &  & &  $ 38.4\% $   \\
        NN-Vit.~\cite{richard2018nnviterbi}   & &  & &  $ 43.0\% $   \\
        \cmidrule(lr){1-5}           
        & \multicolumn{4}{c}{Unsupervised}\\
        \cmidrule(lr){1-5}           
        & & F1-score  & & MoF  \\
        \cmidrule(lr){1-5}           
        Mallow \cite{sener2018unsupervised}   & &  $ - $   & &  $ 34.6\% $     \\
        \textit{Ours}   & &  $ 26.4\% $   & &  $ 41.8\% $     \\
        \bottomrule
    \end{tabularx}
    \vspace{2mm}
    \caption{Comparison of proposed method to other state-of-the-art approaches for fully, weakly and unsupervised learning on the Breakfast dataset. } 
    \label{tab:eval_sota_bf}
\end{table}
We further compare the proposed approach to current state-of-the-art approaches, considering unsupervised learning setups as well as weakly and fully supervised approaches for both datasets.
However, even though evaluation metrics are directly comparable to weakly and fully supervised approaches, one needs to consider that the results of the unsupervised learning are reported with respect to an optimal assignment of clusters to ground-truth classes and therefore report the best possible scenario for this task.

We compare our approach to recent works on the Breakfast dataset in Table~\ref{tab:eval_sota_bf}.
As already discussed in Sec.~\ref{sec:eval_inference}, our approach outperforms the current state-of-the-art for unsupervised learning on this dataset by $7.2\%$. But it also shows that the resulting segmentation is comparable to the results gained by the best weakly supervised system so far \cite{richard2018nnviterbi} and outperforms all other recent works in this field. In the case of YouTube Instructions, we compare to the approaches of \cite{alayrac2016unsupervised} and \cite{sener2018unsupervised} for the case of unsupervised learning only. Note that we follow their protocol and report the accuracy of our system without considering background frames. Here, our approach again outperforms both recent methods with respect to Mof as well as F1-score. A qualitative example of the segmentation on both datasets is given in Fig.~\ref{fig:segm_frames}.

 \begin{table}[t] \footnotesize
   \centering
   \begin{tabularx}{0.45\textwidth}{lXcXc}
        \toprule
        \multicolumn{5}{c}{YouTube Instructions}\\
        \cmidrule(lr){1-5}           
        & \multicolumn{4}{c}{Unsupervised}\\
        \cmidrule(lr){1-5}           
        & & F1-score & & MoF  \\
        \cmidrule(lr){1-5}           
        Frank-Wolfe \cite{alayrac2016unsupervised}  & &  $ 24.4\% $     & &  $ - $    \\
        Mallow \cite{sener2018unsupervised}     & &  $ 27.0\% $  & &  $ 27.8\% $    \\
        \textit{Ours}    & &  $ 28.3\% $   & &  $ 39.0\% $    \\
        \bottomrule
    \end{tabularx}
    \vspace{2mm}
    \caption{ Comparison of the proposed method to other state-of-the-art approaches for unsupervised learning on the YouTube Instructions dataset. We report results for a background ratio $\tau$ of $75\%$. Results of F1-score and MoF are reported without background frames as in \cite{alayrac2016unsupervised,sener2018unsupervised}. } 
    \label{tab:eval_sota_yti}
\end{table}
Although we cannot compare with other unsupervised methods on the 50Salads dataset, we compare our approach with the state-of-the-art for weakly and fully supervised learning in Table \ref{tab:sota_salads}. Each video in this dataset has a different order of subactions and includes many repetitions of the subactions. This makes unsupervised learning very difficult compared to weakly or fully supervised learning. Nevertheless, $ 30.2\% $ and $ 35.5\% $ MoF accuracy are still competitive results for an unsupervised method. 
 \begin{table}[t] \footnotesize
   \centering
   \begin{tabularx}{0.45\textwidth}{lXcXc}
        \toprule
        \multicolumn{5}{c}{50Salads}\\
        \cmidrule(lr){1-5}           
        Supervision & & Granularity level & & MoF  \\
        \cmidrule(lr){1-5}          
        Fully supervised \cite{fermuller2018}  & &  eval     & &  $ 88.5\% $    \\
        Unsupervised (\textit{Ours})    & &  eval   & &  $ 35.5\% $    \\
        \cmidrule(lr){1-5}          
        Fully supervised \cite{ding2017tricor}  & &  mid     & &  $ 67.5\% $    \\
        Weakly supervised \cite{richard2018nnviterbi}    & &  mid & &  $ 49.4\% $    \\
        Unsupervised (\textit{Ours})    & &  mid   & &  $ 30.2\% $    \\
        \bottomrule
    \end{tabularx}
    \vspace{2mm}
    \caption{Comparison of proposed method to other state-of-the-art approaches for fully, weakly and unsupervised learning on the 50Salads dataset. } 
    \label{tab:sota_salads}
\end{table}

\subsection{Unknown Activity Classes}
\label{sec:eval_unsupervised}
Finally, we assess the performance of our approach with respect to a complete unsupervised setting as described in Sec.~\ref{sec:Unsupervised_unknown_activities}. Thus, no activity classes are given and all videos are processed together. For the evaluation, we again perform matching by the Hungarian method and match all subactions independent of their video cluster to all possible action labels. In the following, we conduct all experiments on the Breakfast dataset and report MoF accuracy unless stated otherwise. We assume in case of Breakfast $K'=10$ activity clusters with $K=5$ subactions per cluster, we then match 50 different subaction clusters to 48 ground-truth subaction classes, whereas the frames of the leftover clusters are set to background. For the evaluation of the activity clusters, we perform the Hungarian matching on activity level as described earlier. 
\noindent \textbf{Activity-level clustering.} We first evaluate the correctness of the resulting activity clusters with respect to the proposed bag-of-words clustering. We therefore evaluate the accuracy of the completely unsupervised pipeline with and without bag-of-words clustering, as well as the case of hard and soft assignment. As can be seen in Table \ref{tab:eval_bow_clustering}, omitting the quantization step significantly reduces the overall accuracy of the video-based clustering.
 \begin{table}[t]  \footnotesize
   \centering
   \begin{tabularx}{0.45\textwidth}{lXc}
        \toprule
        \multicolumn{3}{c}{ Accuracy of activity clustering}\\
        \cmidrule(lr){1-3}           
        & &  mean over videos   \\
        \cmidrule(lr){1-3}           
        \textit{no BoW}    & & $  19.3\% $  \\
        \textit{BoW hard ass.}   & & $ 29.8\% $  \\
        \textit{BoW soft ass.}   & & $ 31.8\% $  \\
        \bottomrule
    \end{tabularx}
    \vspace{2mm}
    \caption{ Evaluation of activity based clustering on Breakfast with $K'=10$ activity clusters.  } 
    \label{tab:eval_bow_clustering}
\end{table}

\noindent \textbf{Influence of additional embedding.} We also evaluate the impact of learning only one embedding for the entire dataset as in Fig.~\ref{fig:global} or learning additional embeddings for each video set. The results in Table~\ref{tab:eval_dim_epochs} show that a single embedding learned on the entire dataset achieves 18.3\% MoF accuracy. If we learn additional embeddings for each of the $K'$ video clusters, the accuracy even slightly drops.     
For completeness, we also compare our approach to a very simple baseline, which uses k-Means clustering with $50$ clusters using the video features without any embedding. This baseline achieves only $6.1\%$ MoF accuracy. This shows that a single embedding learned on the entire dataset performs best. 
 \begin{table}[t]  \footnotesize
   \centering
   \begin{tabularx}{0.45\textwidth}{lXc}
        \toprule
        \multicolumn{3}{c}{Multiple embeddings}\\
        \cmidrule(lr){1-3}           
        & & MoF \\
        \cmidrule(lr){1-3}           
        \textit{full w add. cluster emb.}    & &  $ 16.4\% $  \\
        \textit{full w/o add. cluster emb.}    & &  $ 18.3\% $  \\
        \bottomrule
    \end{tabularx}
    \vspace{2mm}
    \caption{ Evaluation of the impact of learning additional embeddings for each video cluster on the Breakfast dataset. } 
    \label{tab:eval_dim_epochs}
\end{table}
\noindent \textbf{Influence of cluster size.} For all previous evaluations, we approximated the cluster sizes based on the ground-truth number of classes. We therefore evaluate how the overall ratio of activity and subaction clusters influences the overall performance. To this end, we fix the overall number of final subaction clusters to 50 to allow mapping to the 48 ground-truth subaction classes and vary the ratio of activity ($K'$) to subaction ($K$) clusters. Table \ref{tab:eval_Influence_cluster_size} shows the influence of the various cluster sizes. It shows that omitting the activity clustering ($K'=1$), leads to significantly worse results. Depending on the measure, good results are achieved for $K'=5$ and $K'=10$.  
 \begin{table}[t]  \footnotesize
   \centering
   \begin{tabularx}{0.45\textwidth}{lXcXcXc}
        \toprule
        \multicolumn{7}{c}{Influence of cluster size}\\
        \cmidrule(lr){1-7}           
        $K'$ / $K$  & &  mean over videos  & &  MoF & & IoU \\
        \cmidrule(lr){1-7}           
        \textit{1 / 50}    & &  $ 10.9\% $  & &  $  10.7\%$ & & $ 4.0\% $ \\
        \textit{2 / 25}    & & $  19.9\%  $   & &  $  15.3\%$ & & $ 5.6\% $ \\
        \textit{3 / 16}    & & $  25.6\%  $  & &   $  16.2\%$ & & $ 6.1\% $ \\
        \textit{5 / 10}    & & $  30.6\%  $   & &  $  18.8\%$ & & $ 7.1\% $ \\
        \textit{10 / 5}    & & $  31.8\%  $   & &  $  18.3\%$ & & $ 13.2\% $ \\
        \bottomrule
    \end{tabularx}
    \vspace{2mm}
    \caption{Evaluation of the number of activity clusters ($K'$) with respect to the number of subaction clusters ($K$) on the Breakfast dataset. The second column (mean over videos) reports the accuracy of the activity clusters ($K'$) as in Table~\ref{tab:eval_bow_clustering}.} 
    \label{tab:eval_Influence_cluster_size}
\end{table}
\noindent \textbf{Unsupervised learning on YouTube Instructions.} Finally, we evaluate the accuracy for the completely unsupervised learning setting on the YouTube Instructions dataset in Table~\ref{tab:eval_background_YTI}. We use $K=9$ and $K'=5$ and follow the protocol described in Sec.~\ref{sec:eval_background}, \ie, we report the accuracy with respect to the parameter $\tau$ as MoF and IoU with and without background frames. As we already observed in Fig.~\ref{fig:eval_background_YTI}, IoU with background frames is the only reliable measure in this case since the other measures are optimized by declaring all or none of the frames as background.      
Overall we observe a good trade-off between background and class labels for $\tau = 75\%$.
 \begin{table}[t]  \footnotesize
   \centering
   \begin{tabularx}{0.45\textwidth}{lXcXcXcXc}
        \toprule
        \multicolumn{9}{c}{Influence of background ratio $\tau$}\\
        \cmidrule(lr){1-9}           
        & & & MoF & & & & IoU &\\
        \cmidrule(lr){1-9}           
        $\tau$ & & wo bg. & & w bg. & & wo bg. & & w bg.\\
        \cmidrule(lr){1-9}           
        \textit{60}    & &  $ 19.8\% $  & & $ 8.0\%$  & & $ 4.9\%$  & & $ 4.9\%$ \\
        \textit{70}    & &  $ 19.6\% $  & & $ 9.0\%$  & & $ 4.9\%$  & & $ 4.8\%$\\
        \textit{75}    & &  $ 19.4\% $  & & $ 10.1\%$  & & $ 4.8\%$  & & $ 4.8\%$\\
        \textit{80}    & &  $ 18.9\% $  & & $ 12.0\%$  & & $ 4.8\%$  & & $ 4.9\%$ \\
        \textit{90}    & &  $ 15.6\% $  & & $ 22.7\%$  & & $ 4.3\%$  & & $ 4.7\%$ \\
        \textit{99}    & &  $ 2.5\% $  & & $ 58.6\%$  & & $ 1.5\%$  & & $ 2.7\%$ \\
        \bottomrule
    \end{tabularx}
    \vspace{2mm}
    \caption{ Evaluation of $\tau$ reported as MoF and IoU without and with background on the YouTube Instructions dataset. } 
    \label{tab:eval_background_YTI}
\end{table}
%
\section{Conclusion}
We proposed a new method for the unsupervised learning of actions in sequential video data. Given the idea that actions are not performed in an arbitrary order and thus bound to their temporal location in a sequence, we propose a continuous temporal embedding to enforce clusters at similar temporal stages. We combine the temporal embedding with a frame-to-cluster assignment based on Viterbi decoding which outperforms all other approaches in the field. Additionally, we introduced the task of unsupervised learning without any given activity classes, which is not addressed by any other method in the field so far. We show that the proposed approach also works on this less restricted, but more realistic task.  

\textbf{Acknowledgment.}
The work has been funded by the Deutsche Forschungsgemeinschaft (DFG, German Research Foundation) – GA 1927/4-1 (FOR 2535 Anticipating Human Behavior), KU 3396/2-1 (Hierarchical Models for Action Recognition and Analysis in Video Data), and the ERC Starting Grant ARCA (677650). This work has been supported by the AWS Cloud Credits for Research program.

{\small
\bibliographystyle{ieee_fullname}
\bibliography{egbib}
}

\end{document}